\pdfoutput=1
\documentclass[11pt,a4paper]{article}
\usepackage[hyperref]{emnlp2020}
\usepackage{times}
\usepackage{latexsym}
\usepackage{tabularx}
\usepackage{graphicx}
\usepackage{adjustbox}
\usepackage{multirow}
\usepackage{booktabs}
\usepackage{amsmath}
\usepackage{cleveref}
\usepackage{todonotes}
\usepackage{float}
\usepackage{nameref}
\usepackage[utf8]{inputenc}

\usepackage[linesnumbered,ruled,vlined]{algorithm2e}
\usepackage{amssymb}
\usepackage{pifont}
\usepackage{enumitem}
\usepackage{textcomp}

\newcommand{\cmark}{\ding{51}}%
\definecolor{greenrgb}{rgb}{0.18, 0.71, 0.18}
\definecolor{purple_corefi}{rgb}{199,21,133}
\definecolor{blue_corefi}{RGB}{45,156,219}
\newcommand\corefi{\textsc{CoRefi}}

\usepackage{microtype}

\aclfinalcopy 

\title{{C}o{R}efi: {A} {C}rowd {S}ourcing {S}uite for {C}oreference {A}nnotation
}

\author{Aaron Bornstein\textsuperscript{1,2} \quad
    Arie Cattan\textsuperscript{1} \quad
    Ido Dagan\textsuperscript{1}\\
\textsuperscript{1}Computer Science Department, Bar Ilan University, Ramat-Gan, Israel\\
\textsuperscript{2}Microsoft Corporation, Tel-Aviv, Israel 
 \\
  {\tt  abornst@microsoft.com arie.cattan@gmail.com dagan@cs.biu.ac.il}
  }

\begin{document}

\maketitle

\begin{abstract}

Coreference annotation is an important, yet expensive and time consuming, task, which often involved expert annotators trained on complex decision guidelines. To enable cheaper and more efficient annotation, we present \corefi{}, a web-based coreference annotation suite, oriented for crowdsourcing. Beyond the core coreference annotation tool, \corefi{} provides guided onboarding for the task as well as a novel algorithm for a reviewing phase. \corefi{} is open source and directly embeds into any website, including popular crowdsourcing platforms.

\corefi{} Demo: \url{aka.ms/corefi}
Video Tour: \url{aka.ms/corefivideo}
Github Repo: \url{https://github.com/aribornstein/corefi}

\end{abstract}

\section{Introduction}

Coreference resolution is the task of clustering textual expressions (\textit{mentions}) that refer to the same concept in a described scenario. This challenging task has been mostly investigated within a single document scope, seeing great research progress in recent years. The rather under-explored cross-document coreference setting is even more challenging. For example, consider the following sentences originating in two different documents in the standard cross-document coreference dataset ECB+ \cite{cybulska-vossen-2014-using}:
\begin{enumerate}[leftmargin=*]
    \item \emph{A man suspected of \textbf{shooting} three people at an \underline{accounting firm} where he had worked ...}
    \item \emph{A gunman \textbf{shot} three people at a suburban \underline{Detroit office building} Monday morning.}
\end{enumerate}
Recognizing that both sentences refer to the same event (``shooting",``shot") at the same location (``accounting firm", ``Detroit office") can be very useful for downstream tasks, particularly across documents, such as multi-document summarization~\cite{falke-etal-2017-concept,liao-etal-2018-abstract} or multi-hop question answering \cite{dhingra-etal-2018-neural, wang-etal-2019-multi-hop}.

High-quality annotated datasets are valuable to develop efficient models. While Ontonotes~\cite{pradhan-etal-2012-conll} provides a useful dataset for generic single-document coreference resolution, large-scale datasets are lacking for cross-document coreference \cite{cybulska-vossen-2014-using, minard-etal-2016-meantime, vossen-etal-2018-dont} or for targeted domains, such as medical~\cite{nguyen-etal-2011-overview}.
Due to the complexity of the coreference task, existing datasets have been annotated mostly by linguistic experts, incurring high costs and limiting annotation scale. 

\begin{figure}[t]
    \centering
    \includegraphics[width=8cm]{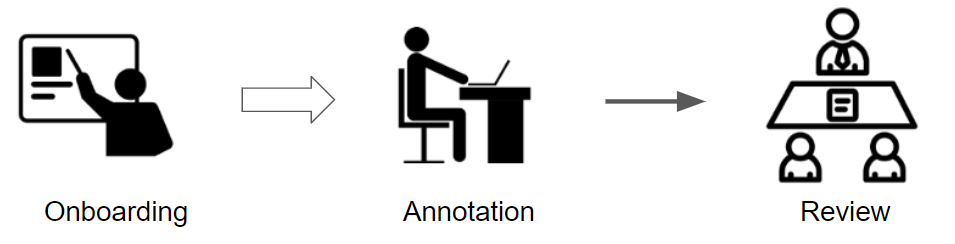}
    \caption{\corefi{}'s end-to-end annotation process}
    \label{fig:e2e}
\end{figure} 

Aiming to address the cost and scalability issues in coreference annotation, we present \corefi{}, an embeddable web-component tool suite that supports an end-to-end crowdsourcing process (Figure~\ref{fig:e2e}), while providing several contributions over earlier annotation tools (Section \ref{sec:related}). \corefi{} includes an automated onboarding training phase, familiarizing annotators with the tool functionality and the task decision guidelines. Then, actual annotation is performed through a simple-to-use and efficient interface, providing quick keyboard operations. Notably, \corefi{} provides a reviewing mode, by which an additional annotator reviews and improves the output of an earlier annotation. This mode is enabled by a non-trivial algorithm, that seamlessly integrates reviewing of an earlier annotation into the progressive construction of the reviewer's annotation. 

By open sourcing \corefi{}, we hope to facilitate the creation of large-scale coreference datasets, especially for the cross-document setting, at modest cost while maintaining quality.

\section{The \corefi{} Annotation Tool}
\label{sec:tool}

\corefi{} provides a suite for annotating single and cross-document coreference, designed to embed into crowdsourcing environments. Since coreference annotation is an involved and complex task, we target a \textit{controlled crowdsourcing} setup, as proposed by \citet{roit-etal-2020-controlled}. This setup consists of selecting designated promising crowd workers, identified in preliminary trap-tasks, and then quickly training them for the target task and testing their performance. This yields a pool of reliable lightly trained annotators, who perform the actual annotation of the dataset.

\corefi{} supports both the annotator training (onboarding) and annotation production phases, as illustrated in Figure \ref{fig:e2e}. The training phase (Section~\ref{subsec:onboarding}) consists of two crowdsourcing tasks, first  teaching the tool's functionality and then practicing guided annotation,  interactively learning basics of the annotation guidelines. The annotation production phase also consists of two crowdsourcing tasks: first-round coreference annotation,  providing a user-friendly interface designed to reduce annotation time 
(Section~\ref{subsec:annotation_mode}), and a novel reviewing task, in which an additional annotator reviews and improves the initial annotation (Section~\ref{subsec:reviewing}).

\subsection{Design Choices}

Our first major design choice regards the annotation flow.
As elaborated in Section \ref{sec:related}, two different coreference annotation flows were prominent in prior work. The local pair-based approach aims at annotation simplicity, often motivated by a crowdsourcing setting. Here, an annotator has to decide for a pair of mentions whether they corefer or not, or to proactively find such pairs of corefering mentions.
Since coreference is annotated at the level mention pairs, it might require, in the worst case, comparing a mention to all other mentions in the text.

In the cluster-based flow, annotators assign mentions to coreference \textit{clusters}. Here, a mention needs to be compared only against the clusters accumulated so far, or otherwise be defined as starting a new cluster. Indeed, the number of coreference clusters is often substantially lower than the number of mentions, particularly in the cross-document setting, where the same content gets repeated across the multiple texts. For example, in the most popular dataset for cross-document coreference, ECB+~\cite{cybulska-vossen-2014-using}, the number of clusters is about one third of the number of mentions (15122 mentions split into 4965 coreference clusters, including singletons).
In \corefi{}, we adopt the cluster-based approach since we aim at exhaustive coreference annotation across documents, whose complexity would become too high under the pairwise approach. At the same time, we simplify the annotation process and functionality, making it crowdsourceable.

Our second design choice regards detecting referring mentions in text. As elaborated in Section \ref{sec:related}, coreference annotation tools, particularly cluster-based (e.g. \cite{Reiter2018ag,oberle-2018-sacr}), 
often require annotators to first detect the target mentions before annotating them for coreference.
Conversely, recent local pair-based decision tools \cite{chamberlain-etal-2016-phrase, li-etal-2020-active} delegate mention extraction to a preprocessing phase, presenting coreference annotators with pre-determined mentions. This simplifies the task and allows annotators to focus their attention on the coreference decisions. 

As we target exhaustive crowdsourced coreference annotation, we chose to follow this recent facilitating approach. In addition to the input texts, \corefi{} takes as input an annotation of the targeted mentions, while optionally allowing annotators to fix this mention annotation. 
In our tool suite, we followed the approach of Prodigy,\footnote{\url{https://prodi.gy/}} where corpus developers may implement their own automated (non-overlapping) mention extraction recipes, or use a separate manual annotation tool for mention annotation, according to their desired mention detection guidelines (which often vary across projects). The resulting mentions can then be fed into \corefi{} for coreference annotation. We provide an example mention extraction recipe that detects as mentions common nouns, proper nouns, pronouns, and verbs (for event coreference). Such mention detection is consistent with approaches that consider reduced mention spans, mostly pertaining to syntactic heads or named entity spans \cite{ogorman-etal-2016-richer}.

\subsection{Annotation}
\label{subsec:annotation_mode}

\begin{figure*}[!ht]
    \centering
    \frame{\includegraphics[width=\textwidth]{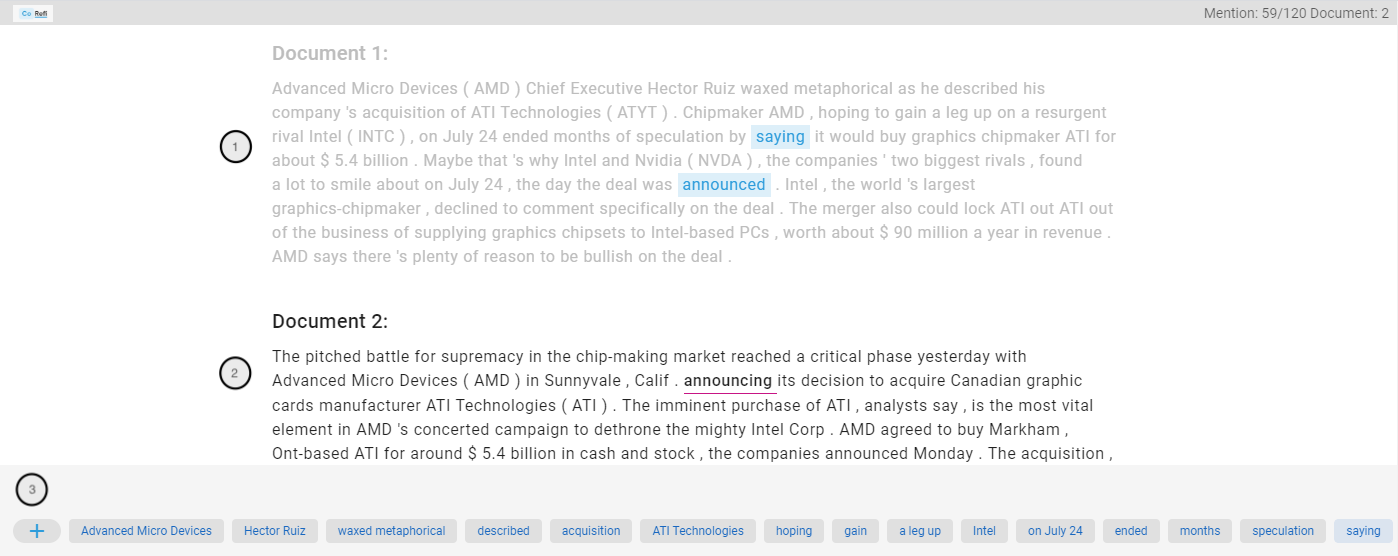}}
    \caption{Annotation Interface of \corefi{}, presenting text and mentions from the ECB+ dataset, used in our pilot study (Section \ref{sec:study}).
    The current mention to assign is underlined in purple (2). The selected cluster is highlighted in blue in the cluster bank (3), along with its mentions in the text (1).
    }
    \label{fig:corefi_annotation_mode}
\end{figure*}

Figure~\ref{fig:corefi_annotation_mode} shows the annotation interface of \corefi{}.

As initialization, the first candidate mention is automatically assigned to the first coreference cluster, which is placed in the ``cluster bank", appearing at the bottom of the screen ((3) in Figure~\ref{fig:corefi_annotation_mode}). In this bank, each cluster is labeled by the text of its first mention.
The annotator is then shown the subsequent mentions, one at a time, with the current mention to assign underlined in purple~(2).  
For each mention, the annotator decides whether to accept it as a valid mention (doing nothing) or to modify its span (easily highlighting the correct span and pressing the `F' key, for ``Fix"). Similarly, the annotator may introduce a new span, missing from the input. To simplify annotation, the tool allows only non-overlapping spans.

The annotator then makes a coreference decision, by assigning the current mention to a new or existing cluster.
An existing cluster can be rapidly selected either by selecting it in the cluster bank or by selecting one of its previously-assigned mentions in the text. Once a cluster is selected, it is highlighted in blue along with all its previous text mentions ((3) and (1) in the figure).
Rather than assigning mentions to clusters through a slower drag and drop interface \cite{Reiter2018ag, oberle-2018-sacr} or buttons \cite{girardi-etal-2014-cromer, aralikatte-sogaard-2020-model}, annotation is driven primarily by faster keyboard operations, such as \textsc{Space} (assign to an existing cluster) and \textsc{Ctrl+Space} (new cluster), with quick navigation through arrow keys and mouse clicks.

At any point, the annotator can re-assign a previously assigned mention to another cluster 
or view any cluster mentions. 
\corefi{} supports an unlimited number of documents to be annotated, presented sequentially in a configurable order. 
Finally, \corefi{} guarantees exhaustive annotation by allowing task submission only once all candidate mentions are processed.

\subsection{Reviewing}
\label{subsec:reviewing}

To promote annotation quality, annotation projects typically rely on multiple annotations per item. One approach for doing that involves collecting such annotations in parallel and then merging them in some way, such as simple or sophisticated voting \cite{hovy-etal-2013-learning}. 
Another approach is sequential, where one or more annotations are collected initially, and are then \textit{manually} consolidated by an additional, possibly more reliable, annotator (a ``consolidator" or ``reviewer") \cite{roit-etal-2020-controlled}.

In our case, coreference annotation is addressed as a global clustering task, where an annotator generates a complete clustering configuration for the input text(s). 
Automatically merging such multiple clustering configurations, where cluster assignments are mutually dependent, might become unreliable. 
Therefore, in \corefi{} we follow the sequential manual reviewing approach. To that end, we introduce a novel reviewing task, which receives as input a previously annotated clustering configuration and allows an additional annotator to review and improve it.

\begin{figure*}[!ht]
    \centering
    \frame{\includegraphics[width=\textwidth]{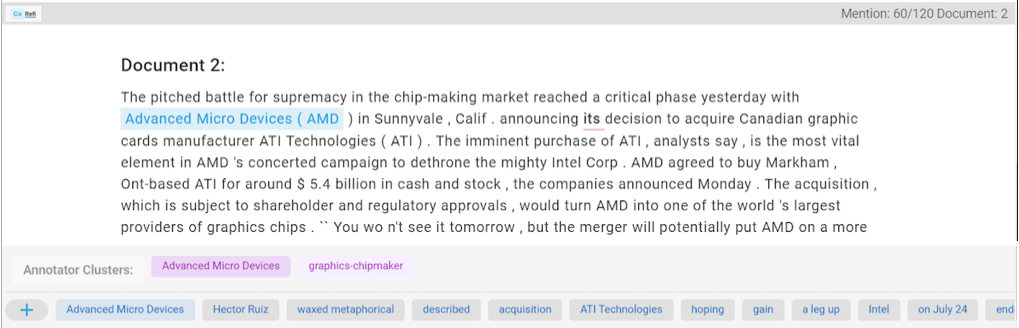}}
    \caption{Reviewing interface of \corefi{}. Candidate clusters found by the reviewing algorithm are shown in purple.}
    \label{fig:corefi_review_mode}
\end{figure*}

The reviewing task follows the same flow of the annotation task, making it trivial to learn for annotators that already experienced with \corefi{} annotation. At each step, the reviewer is presented with the next mention in the reviewed configuration, and may first decide to modify its span. Next, the reviewer has to decide on cluster assignment for the current mention. The only difference at this point is that the reviewer is presented with candidate cluster assignments which reflect the original annotator assignment (as explained below), displayed just above the cluster bank (Figure~\ref{fig:corefi_review_mode}).

In fact, it is not trivial to reflect the cluster assignment by the original annotator to the reviewer, since that assignment has to be mapped to the current clustering configuration of the reviewer. Ambiguity may arise, resulting in multiple candidate clusters, since an early cluster modification by the reviewer can impact the interpretation of downstream cluster assignments in the original annotation.

To illustrate this issue, consider reviewing a cluster assigned by the original annotator, consisting of three mentions, $\{A, B, C\}$. 
When presented with the mention $A$, the reviewer agrees that it starts a new cluster. Then, when reaching $B$, the reviewer is presented with $\{A\}$ as $B$'s original cluster assignment. Suppose the reviewer disagrees with the annotator that $A$ and $B$ corefer and decides to assign $B$ to a new cluster. Now, when reviewing the mention $C$, it is no longer clear whether to attribute $C$'s original assignment to $\{A\}$ or $\{B\}$. Hence, the reviewing tool presents both $\{A\}$ and $\{B\}$ as candidate clusters that reflect the original annotator's assignment. The reviewer may then choose either of them, or override the original annotation altogether and make a different assignment. Similar ambiguities arise when the reviewer splits an original mention span and assigns its parts to separate clusters.

To address this challenge, we formulate an algorithm that maps an original cluster assignment to a set of candidate clusters in the current clustering configuration of the reviewer. Generally speaking, the algorithm considers the cluster to which the current mention was assigned in the original annotation, and tracks all earlier token positions in that cluster. These token positions are then mapped back to clusters in the current reviewer's clustering configuration, which become the candidate clusters presented to the reviewer. The algorithm pseudocode is presented in Appendix~\ref{app:review_algo_pseudocode}, along with a comprehensive example of its application.

\subsection{Onboarding}
\label{subsec:onboarding}

In the proposed controlled crowdsourcing scheme of  \newcite{roit-etal-2020-controlled}, annotators were trained in an ``offline" manner, reading slides and receiving individual feedback. We propose augmenting this phase with automated training, delivered through two crowdsourcing tasks, described below.

\subsubsection{Walk-through Tutorial}
\label{subsubsec:tutorial}

During this task, a trainee is walked through the core concepts and functionality of \corefi{}, such as the ``current mention" and the ``cluster bank" (Figure~\ref{fig:corefi_annotation_mode}), and through the annotation operations.
These functionalities are presented through a series of intuitive dialogues. To ensure that each feature is correctly understood, the user is instructed to actively perform each operation before continuing to the next (see Appendix~\ref{app:turorial}).

\subsubsection{Guided Annotation}
\label{subsubsec:guided_hit}

After acclimating with \corefi{}'s features, users are familiarized with the coreference decision guidelines through a guided annotation task, practicing annotation while receiving automated guiding feedback. If an annotation error is made, the trainee is notified with a pre-prepared custom response, which guides to the correct decision before allowing to proceed.  Additionally, following certain decisions, specific important guidelines can be communicated (see Appendix~\ref{app:guided} for examples).

The content of the guided annotation task and the automated responses are easily configurable using a simple JSON configuration schema. This allows tailoring them  when applying \corefi{} for different datasets and annotation guidelines.

Augmenting the controlled crowdsourcing scheme \cite{roit-etal-2020-controlled} with automated training provides key  benefits. 
First, since feedback is automated, the  amount of required personalized manual feedback is reduced. 
Second, annotators benefit from an immediate response for each decision, allowing them to understand their mistakes earlier  and improve in real time.
We suggest that, for optimal learning of annotation guidelines, these benefits should be coupled with the additional training means of controlled crowdsourcing. These include the provision of guideline slides, for learning and later for reference, and some personalized manual feedback during the training phase.

\subsection{Implementation Benefits}

\corefi{} was developed using the web component standard and the \textsc{Vue.js} framework.\footnote{\url{https://vuejs.org/}} allowing, it to easily embed into any website, including crowdsourcing platforms. Additionally, \corefi{} provides output in the standard CoNLL coreference annotation format, enabling training state of the art models and scoring with the official coreference scorer \cite{pradhan-etal-2014-scoring}. All \corefi{} features are easily configurable with HTML encoded JSON and support any UTF-8 encoded language.

\section{Pilot Study}
\label{sec:study}

To further assess \corefi{}'s effectiveness in a crowdsourcing environment, we performed a small-scale trial on Amazon Mechanical Turk, employing 5 annotators, focusing on the coreference annotation functionality (rather than mention validation).
To allow objective assessment of annotation quality, we experimented with replicating coreference annotations from the ECB+ dataset \cite{cybulska-vossen-2014-using}, the commonly used dataset for cross-document coreference over English news articles \cite{cybulska-vossen-2015-translating, yang-etal-2015-hierarchical, choubey-huang-2017-event, kenyon-dean-etal-2018-resolving, barhom-etal-2019-revisiting, Cattan2020StreamliningCC}.
Accordingly, we considered the ECB+ gold mentions as input, requesting crowdworkers to assign them to coreference clusters.
Focusing on the controlled crowdsourcing setting, we hired five annotators that were previously selected for annotation by \citet{roit-etal-2020-controlled}.

\paragraph{On-boarding}

Annotators were given \corefi{}'s walk-through tutorial and guided annotation tasks, adapted to the ECB+ guidelines and
applied to a part of an ECB+ subtopic (cluster of documents).
These two tasks took altogether 11 minutes on average to complete, at a rate of \$1.5 a task. 
Next, workers were asked to annotate an entire ECB+ subtopic (through the actual annotation task). We provided them manual feedback for their mistakes, which consumed 30-40 minutes of researcher time per trained annotator.

\paragraph{Annotation}

After training, we paid workers \$8 to annotate two additional subtopics in full (of about 150 and 200 mentions; in ECB+ only a few sentences are annotated per document, and these were presented for annotation). Each subtopic took 27 minutes on average to annotate, corresponding to an annotation rate of \texttildelow{}400 mentions per hour. 

Table~\ref{tab:annotation_results} presents the performance (F1) of each of the annotators, compared to the ECB+ gold annotations, averaged over the two subtopics. The results are reported using the common evaluation metrics for coreference resolution: \textbf{MUC} \cite{vilain-etal-1995-model}, \textbf{B\textsuperscript{3}} \cite{bagga-baldwin-1998-entity-based}, \textbf{CEAFe} \cite{luo-2005-coreference}, and \textbf{CoNLL} --- the average of the three metrics.
Considering the decision volume and complexity, as well as the limited training (not providing guideline slides and a single practice round), we find that these results support \corefi{}'s effectiveness for crowdsourcing.\footnote{There are no comparable annotator performance evaluations in the literature. Ontonotes~\cite{pradhan-etal-2012-conll} reports an inter-annotator agreement for experts of 0.87 MUC scores, but these seem to include mention span decisions. The creators of ECB+~\cite{cybulska-vossen-2014-using} use a different methodology to calculate inter-annotator agreement, not applicable for our setting, reporting a Kappa score of 0.76.} 
As previously mentioned, we expect that annotation quality may be further improved in an actual dataset creation project, by providing additional guided tasks, annotation guidelines slides, and additional manual feedback.

\begin{table}[t]
    \centering
    \resizebox{0.48\textwidth}{!}{
    \begin{tabular}{@{}lcccccccc@{}}
    \toprule
    & \phantom{ab} & \textbf{MUC} & \phantom{a} & \textbf{B\textsuperscript{3}} & \phantom{a}
    & \textbf{CEAFe} & \phantom{a} & \textbf{CoNLL} \\
    \midrule
    A1 && 94.0 && 85.0 && 77.8 && 85.6\\
    A2 && 94.8 && 91.2 && 85.4 && 90.5 \\
    A3 && 95.2 && 90.2 && 84.7 && 90.0 \\
    A4 && 94.8 && 86.9 && 76.0 && 85.9\\
    A5 && 92.1 && 82.5 && 75.7 && 83.4\\
    \bottomrule
    \end{tabular}}
    \caption{F1 results of 5 annotators on 2 ECB+ subtopics.}
    \label{tab:annotation_results}
\end{table}

\paragraph{Reviewing}
For the reviewing trial, the two best annotators, A2 and A3, were selected as R1 and R2. Two additional annotators (A1 and A5) were each assigned a new unique subtopic, which was then reviewed by both R1 and R2. 
Table~\ref{tab:review_results} presents the reviewing results, showing consistent improvements after reviewing and assessing the ease of using the reviewing functionality.

\begin{table}[t]
    \centering
    \resizebox{0.48\textwidth}{!}{
    \begin{tabular}{@{}lcccccccc@{}}
    \toprule
    & \phantom{ab} & \textbf{MUC} & \phantom{a} & \textbf{B\textsuperscript{3}} & \phantom{a}
    & \textbf{CEAFe} & \phantom{a} & \textbf{CoNLL} \\
    \midrule
    A1 && 87.0 && 69.1 && 62.6 && 72.9 \\
    R1 && 91.9 && 80.4 && 79.8 && 80.0\\
    R2 && 88.0 && 73.4 && 73.3 && 78.2 \\
    \midrule
    A5 && 87.0 && 79.0 && 62.5 && 76.2\\
    R1 && 92.9 && 87.3 && 73.0 && 84.4\\
    R2 && 90.0 && 86.2 && 63.0 && 79.7\\
    \bottomrule
    \end{tabular}}
    \caption{F1 results of the reviewing trial.}
    \label{tab:review_results}
\end{table}

\section{Related work}
\label{sec:related}

As mentioned in Section \ref{sec:tool}, prior tools for coreference annotation are based on two prominent workflows: pair-based, treating coreference as a pairwise annotation decision, and cluster-based, in which mentions are assigned to clusters. 
While targeting simplicity, only two pair-based tools supported crowdsourcing annotation, yet they were not applied for producing exhaustively annotated daasets: Phrase Detective \cite{chamberlain-etal-2016-phrase}, which was employed in a web-based game setting, and \cite{li-etal-2020-active}, which was applied in an active learning environment.
 
Pair-based tools differ in their annotation approaches. In certain tools, such as BRAT \cite{stenetorp-etal-2012-brat}, Glozz \cite{10.1145/2361354.2361394}, Analec \cite{landragin-etal-2012-analec}, and MMAX2 \cite{kopec-2014-mmax2},
the annotator first determines mention span boundaries and then links a pair of mentions. Other Pair-based tools \cite{chamberlain-etal-2016-phrase,li-etal-2020-active} either provide annotators a single (pre-determined) mention, asking to find a coreferring antecedent, or provide a pair of mentions, asking to judge whether the two corefer. 
Notably, pair-based tools are less effective for exhaustive coreference annotation, for two reasons. First, they require comparing each mention to all other \emph{mentions}, rather than to already constructed clusters. Second, local pairwise decisions lack awareness of previous cluster assignments, which might hurt annotation quality.

Cluster-based tools, including Cromer~\cite{girardi-etal-2014-cromer}, Model based annotation tool \cite{aralikatte-sogaard-2020-model}, CorefAnnotator~\cite{Reiter2018ag}, and SACR~\cite{oberle-2018-sacr}, ask annotators to first detect mention spans and then cluster them, thus complicating the overall task without allowing the delegation of mention detection to a preprocessing phase.
Such a method does not guarantee exhaustive annotation, since annotators may miss some mentions. With respect to operation efficiency, mentions are often linked to clusters via somewhat slow operations, such as drag-and-drop or selection from a drop-down list, in comparison to the fast keyboard operations in \corefi{}.

Notably, to the best of our knowledge, \corefi{} is the first cluster-based crowdsourcing tool that provides an end-to-end annotation suite, including automated onboarding tasks and exhaustive reviewing, the latter enabled by our novel reviewing algorithm. Furthermore, it is the first tool that was developed using the WebComponent standard, embeddable in any website.

\section{Conclusion}

In this paper, we aim to facilitate crowdsourced creation of needed large-scale coreference datasets, in both the within- and the cross-document setting. 
Our comprehensive end-to-end tool suite, \corefi{}, enables high quality and fairly cheap crowdsourcing of exhaustive coreference annotation in various domains and languages. Our experiments demonstrate that \corefi{}'s automatic onboarding is effective at augmenting \newcite{roit-etal-2020-controlled}'s controlled crowdsourcing. \corefi{} provides the first reviewing algorithm and implementation for cluster-based coreference annotation. Overall, we demonstrated that non-expert annotators can be trained to effectively perform and review coreference annotations, allowing for cost-effective annotation efforts.

\section*{Acknowledgments}

The work described herein was supported in part by grants from Intel Labs, Facebook, the Israel Science Foundation grant 1951/17, the Israeli Ministry of Science and Technology and the German Research Foundation through the German-Israeli Project Cooperation (DIP, grant DA 1600/1-1).

In addition to the support above, we would like to thank Uri Fried, Ayal Klein, Paul Roit, Amir Cohen, Sharon Oren, Chris Noring, Asaf Amrami, Ori Shapira, Daniela Stepanov, Ori Ernst, Yehudit Meged, Valentina Pyatkin, Moshe Uzan, and Ofer Sabo for their support with architecture, design and crowdsourcing.

\bibliographystyle{acl_natbib}
\bibliography{anthology,emnlp2020}

\appendix

\section{Reviewing Algorithm}
\label{app:review_algo_pseudocode}

Algorithm~\ref{review_algo} implements the mechanism to find potential clusters given the initial annotation and previous reviewing modifications.

\definecolor{mygray}{gray}{0.2}
\newcommand\mycommfont[1]{\footnotesize\ttfamily\textcolor{mygray}{#1}}
\SetCommentSty{mycommfont}

\begin{algorithm}
\SetAlgoLined
\DontPrintSemicolon 
\KwIn{$M$: Stack of mentions with their initial clustering assignment}
\KwOut{$R$: Reviewed Assignment} 
\BlankLine

$Ant \gets CreateAntecedentMapping(M)$\;
$T2C$: Map of token to cluster ID\;
 \While{$M$ not empty}{
 \CommentSty{// Set reviewer span}\;
    $\ Sp' \gets$ ReviewSpan($M.top()$)\;
    \While{$M.top().end \leq Sp'.end$}{
    $M.pop()$\;
    }
    \If{$M.top().start \leq Sp'.end$}{
    popSplitPush($M, Sp'$)}
\BlankLine
\CommentSty{// Set reviewer cluster}\;    
  $C \gets getCandidates(Sp', Ant, T2C)$\;
  $cluster \gets selectCluster(C)$\;
  $T2C.update(Sp', cluster)$\;
  $R.push(Sp', cluster)$\;
 }
 \caption{Reviewing Algorithm}
 \label{review_algo}
\end{algorithm}

To support span modification, we build two main data structures at the \emph{token} level (lines 1 and 2). 
Given the original annotation $M$, we build a static mapping $Ant$ (line 1), where each single token is associated with all tokens from previous mentions that belong to the same coreference cluster.
$T2C$ is a growing mapping that will keep track of the reviewer decisions.

After the initialization phase, the reviewer is shown all the annotator mentions in a sequential order. 
For each presented mention, the reviewer first decides whether to agree or to modify the mention span boundaries (line 5).
Future mentions in the stack $M$ that are fully covered by the reviewed span need to be removed (lines 6-8). 
The reviewer may also split the current mention or partially cover next mentions (line 9-11).

In order to find the potential coreference clusters (line 13), we first use $Ant$ to retrieve the antecedent tokens in the original annotation, for each single token in the reviewed span $Sp'$. Then, we use the reviewer mapping $T2C$ for each antecedent tokens to identify the possible cluster(s) that will be displayed to the reviewer (Figure~\ref{fig:corefi_review_mode}). Given the coreference decision (assigning to an existing cluster or to a new one) of the reviewer (line 14), we update the reviewer mapping $T2C$ (line 15) and coreference assignments (line 16). 

Table~\ref{tab:review} illustrates the reviewing decision step by step, given an initial annotation that incorrectly assigned the following gold clustered mentions \{\{Bank of America, bank, BoA\} \{American\}\} into one coreference cluster \{Bank of America, American bank, BoA\}.

\begin{table*}[tb]
    \centering
    \resizebox{\textwidth}{!}{
    \begin{tabular}{llllp{75mm}}
    \toprule
    & \textbf{Mention Stack} & \textbf{Annotator Candidates} & \textbf{Reviewer Decision} 
    & \textbf{Explanation} \\
    \toprule
    1 & [ \textbf{Bank Of America}, American bank, BoA ] & \{Bank of America\} & \textbf{\textcolor{greenrgb}{\cmark}} & The reviewer agrees that Bank of America is the start of a new cluster  \\
    \midrule
    2 & [ \textbf{American bank}, BoA ] & \{Bank of America\} & Split \textbf{American bank} into two mentions & Two mentions are created, \textbf{American} and \textbf{bank}. The reviewer will next determine the cluster assignment of American.
    \\ 
    \midrule
    3 & [ \textbf{American}\, bank, BoA ] & \{Bank of America\} & Assign \textbf{American} to a new cluster & 
    The reviewer is shown \{Bank of America\} as the candidate cluster since the \emph{token} \textbf{American} was assigned with Bank of America by the annotator. \\
    \midrule
    4 & [ \textbf{bank}, BoA ] & \{Bank of America\}, \{American\} & Assign \textbf{bank} to the \{Bank of America\} & 
    The reviewer is shown both \{Bank of America\} and \{American\} as candidate clusters for bank. Now, the reviewer decides to assign bank to the \{Bank of America\} cluster. 
    \\
    \midrule
    5 & [ \textbf{BoA} ] & \{\{Bank of America, bank\}, \{American\}\} & Assign \textbf{BoA} to cluster \{Bank of America, bank\} & The reviewer is shown two candidate clusters \{Bank of America, bank\} and \{American\} which correspond to the clusters that include the antecedent tokens of \{BoA\} initially assigned by the annotator (Bank of America,  American bank, BoA).
    \\
    \bottomrule
    \end{tabular}}
    \caption{Examples of reviewing assignment, the initial clustering assignment is [ (Bank of America, American bank, BoA ] and the reviewer modifies into [ (Bank of America, bank, BoA), (American) ]}
    \label{tab:review}
\end{table*}

\section{Tutorial}
\label{app:turorial}

\begin{figure}[!h]
    \centering
    \frame{\includegraphics[width=0.48\textwidth]{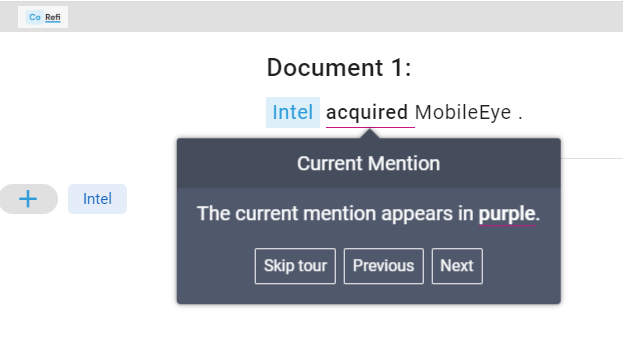}}
    \caption{Example of the tutorial explaining the current mention.}
    \label{fig:tutorial_mention}
\end{figure}
\begin{figure}[!h]
    \centering
    \frame{\includegraphics[width=0.48\textwidth]{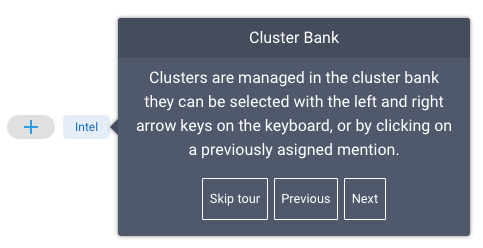}}
    \caption{Example of the tutorial explaining the cluster bank.}
    \label{fig:tutorial_cluster_bank}
\end{figure}
\begin{figure}[!h]
    \centering
    \frame{\includegraphics[width=0.48\textwidth]{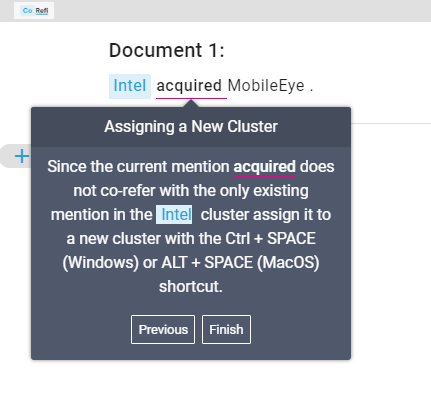}}
    \caption{Example of the tutorial explaining the cluster assignment operation.}
    \label{fig:tutorial_assignment}
\end{figure}

Figure~\ref{fig:tutorial_mention} and~\ref{fig:tutorial_cluster_bank} demonstrate notifications that explains conceptual aspect of \corefi{}. Figure~\ref{fig:tutorial_mention} explains what the current mention to assign is where as Figure~\ref{fig:tutorial_cluster_bank} explains what clusters are and how to manage them in the cluster bank.
Figure~\ref{fig:tutorial_assignment} demonstrates a more interactive tutorial prompt. It shows how to make an active coreference decision with the keyboard and encourages the trainee to experiment with in the confines of the tutorial environment to familiarize themselves with the feature.

\section{Guided Annotation}
\label{app:guided}

\begin{figure*}
    \centering
    \frame{\includegraphics[width=\textwidth]{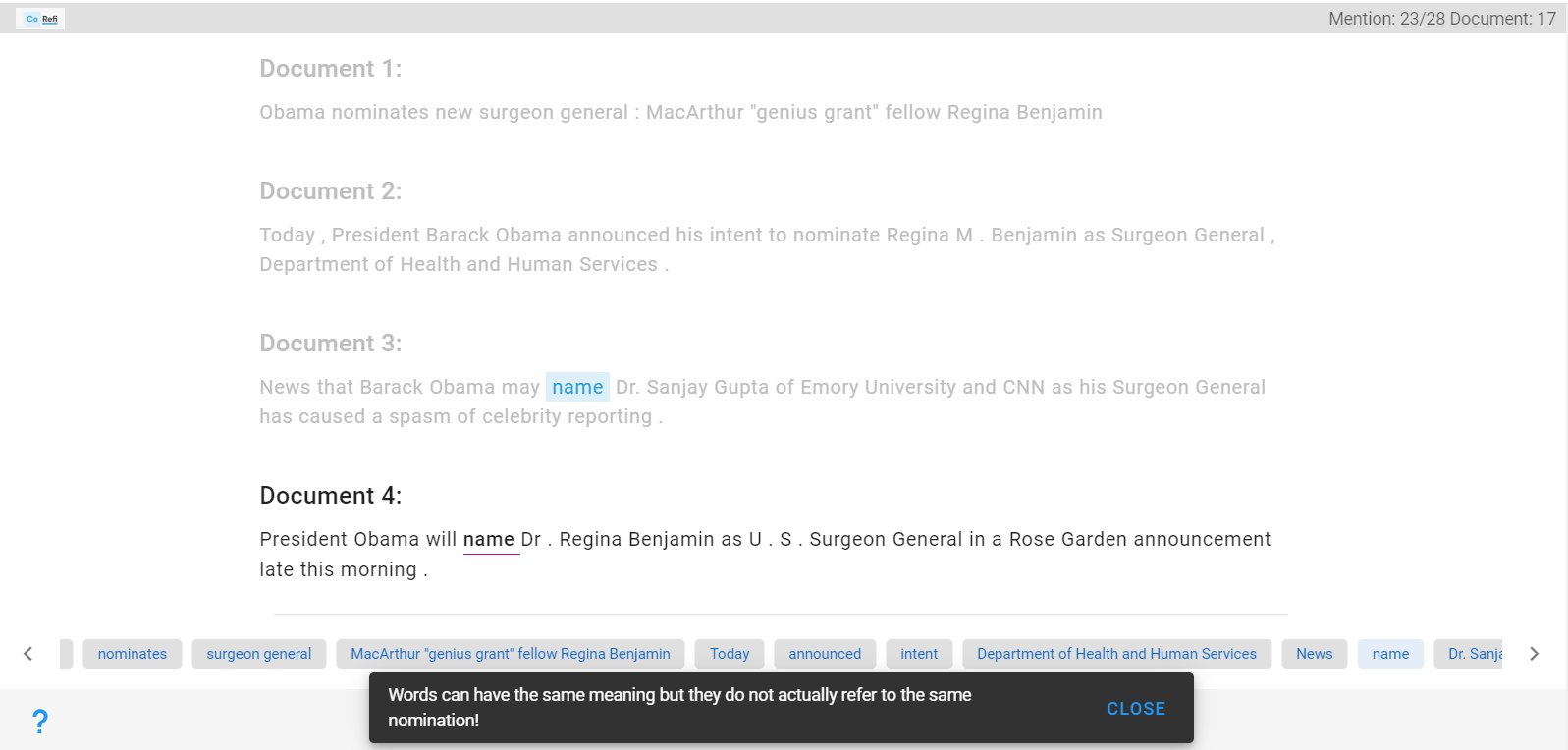}}
    \caption{Example of an automatic feedback during the guided annotation.}
    \label{fig:guided}
\end{figure*}

Figure~\ref{fig:guided} demonstrates the guided experience of the on-boarding flow of \corefi{}. In Figure~\ref{fig:guided}, the trainee is learning the nuances of coreference and makes the mistake of attempting to assign the mention name to the same cluster as another mention with the exact same expression. However, in context the name event mention expressed by the current mention does not refer to the selected cluster. The current mention refers to the naming of the Dr. Regina Benjamin as U.S Surgeon General where as the selected cluster refers to the event of naming Dr. Sanjay Gupta to Surgeon General. Since, the correct decision is subtle the user receives a toast informing them that Words can have the same meaning but not corefer. This toast helps to guide the annotator to the correct decision and reinforces the coreference guidelines. As the trainee is familiarized with the subtleties of coreference they are less likely to make similar mistakes during annotation of the real dataset.

\end{document}